\documentclass[3p,times]{elsarticle}

\usepackage{amssymb}
\usepackage{hyperref}
\usepackage{float}
\usepackage{xcolor}
\usepackage{booktabs}
\usepackage{multirow}
\usepackage{makecell}
\usepackage{comment}

\begin{document}

\begin{frontmatter}

\title{Mawqif-XT: An Arabic Benchmark Dataset for Cross-Target Stance Detection}

\author[1,3]{Rasha Albalawi}
\ead{g202521490@kfupm.edu.sa}
\author[1,2]{Nuha Albadi\corref{cor1}}
\ead{nuha.badi@kfupm.edu.sa}
\author[1,2]{Hamzah Luqman}
\ead{hluqman@kfupm.edu.sa}
\author[1]{Maram Kurdi}
\ead{maram.kurdi@kfupm.edu.sa}
\author[1]{Saad Ezzini}
\ead{saad.ezzini@kfupm.edu.sa}
\author[1]{Asma Yamani}
\ead{g201906630@kfupm.edu.sa}
\author[1]{Ahmed Ashraf}
\ead{g202411740@kfupm.edu.sa}

\cortext[cor1]{Corresponding author}

\address[1]{Information and Computer Science Department, KFUPM, Saudi Arabia}
\address[2]{SDAIA-KFUPM JRC for AI, KFUPM, Saudi Arabia}
\address[3]{University of Tabuk, Saudi Arabia}

\begin{abstract}
Publicly available Arabic datasets for target-specific stance detection remain limited, particularly for evaluating cross-target generalization. This paper presents the Mawqif-XT Extension, consisting of 996 manually annotated Arabic tweets collected from three public targets: Women Driving, E-Cars, and Trimester System. Each tweet is annotated with stance, sentiment, and sarcasm labels following the original Mawqif annotation scheme. The released extension is intended as a held-out evaluation set for assessing model generalization to both semantically related and previously unseen targets, while the original Mawqif dataset is used for training and development. In addition, we establish baseline results using several Arabic and multilingual transformer models, as well as zero-shot large language models (LLMs), to facilitate reproducible evaluation. Together with the original Mawqif dataset, the Mawqif-XT Extension provides a benchmark for evaluating cross-target generalization in Arabic stance detection.
\end{abstract}

\end{frontmatter}

\section{Introduction and Related Work}
Social media platforms have become a major source of user-generated content, where people express opinions on political, social, and public issues. This has increased interest in target-specific stance detection, which aims to identify whether the author is in favor of, against, or neutral toward a given target, even when the target is not explicitly mentioned~\cite{mohammad2016semeval}. The task is particularly challenging for Arabic because of its linguistic diversity and the informal nature of user-generated social media content~\cite{alturayeif2022mawqif}. 

Several Arabic datasets have been introduced for stance detection, covering different data sources, annotation schemes, and evaluation settings. Target-independent resources include the Arabic Fact-Checking corpus~\cite{baly2018integrating}, the Arabic News Stance dataset~\cite{khouja2020stance}, and AraStance~\cite{alhindi2021arastance}, which formulate stance between claims and supporting evidence or news articles. Target-specific resources include Mawqif~\cite{alturayeif2022mawqif}, which provides Arabic tweets annotated for stance, sentiment, and sarcasm. MARASTA~\cite{charfi2024marasta} further extends Arabic stance resources by introducing a multi-dialectal cross-domain corpus covering multiple controversial topics across different Arab regions. Cross-target datasets such as EXaASC~\cite{jaziriyan2021exaasc} and ArabicStanceX~\cite{alkhathlan2025constructing} have expanded evaluation across multiple targets and topics, supporting the study of model generalization.

The original Mawqif dataset introduced the first multi-label benchmark for Arabic stance detection by jointly annotating stance, sentiment, and sarcasm~\cite{alturayeif2022mawqif}. However, because the training and evaluation targets are predefined, it does not enable systematic evaluation of how well models generalize to new targets. This motivates the need for a benchmark specifically designed to evaluate cross-target transfer. Mawqif-XT extends the original dataset with three additional public targets while preserving the original annotation scheme. Women Driving is thematically related to Women Empowerment in the original Mawqif dataset, whereas E-Cars and Trimester System are new targets from different domains. This design enables evaluation on both related-target transfer and unseen-target generalization while preserving the original annotation scheme. This design provides a more comprehensive benchmark for studying cross-target generalization under different transfer scenarios.

This paper presents the Mawqif-XT Extension, which consists of 996 manually annotated Arabic tweets collected from three public targets: Women Driving, E-Cars, and Trimester System. Each tweet is annotated by three annotators with stance, sentiment, and sarcasm labels following the original Mawqif annotation scheme. The dataset is intended as a held-out test set for evaluating cross-target stance detection. The original Mawqif dataset is used for training and development, while the released dataset is used for testing. The test set contains three newly introduced targets. Trimester System and E-Cars evaluate generalization to previously unseen targets, whereas Women Driving is semantically related to Women Empowerment, enabling evaluation on both unseen and related targets. Each tweet is annotated with stance, sentiment, and sarcasm labels, together with the corresponding target and tweet metadata.

%% 5. EXPERIMENTAL DESIGN, MATERIALS AND METHODS
\section{Mawqif-XT Dataset}

The Mawqif-XT Extension extends the original Mawqif dataset while preserving its annotation scheme and overall structure. To ensure compatibility with the original benchmark, the same annotation guidelines, label definitions, and data format were retained. The extension introduces three additional public targets covering different public issues that received considerable attention on social media. Trimester System and E-Cars were selected as previously unseen targets from different domains, whereas Women Driving was chosen because it is closely related to the Women Empowerment target included in the original dataset. This allows the dataset to support evaluation on both a semantically related target and previously unseen targets within a unified evaluation framework. The following subsections describe the stages used to construct the Mawqif-XT Extension.

\subsection{Data Collection}
 The three new targets were collected separately using target-specific search strategies tailored to each topic. Table~\ref{tab:search_strategy} summarizes the search and exclusion terms used for each target.

Arabic tweets related to the \textit{E-Cars} target were collected from X (formerly Twitter) using the Twitter API and a keyword-based search strategy. The search query included multiple Arabic spelling variations of the expressions corresponding to \textit{electric car} and \textit{electric cars}, allowing the collection to capture variations commonly used in Arabic discussions of electric vehicles. The exact keywords used are listed in Table~\ref{tab:search_strategy}. We collected 5,006 tweets covering the period from January 1, 2022 to February 15, 2022.

Arabic tweets related to the \textit{Trimester System} target were collected from X (formerly Twitter) using the Twitter API and a keyword-based search strategy. The query included multiple Arabic lexical and spelling variations corresponding to the following expressions: \textit{three semesters}, \textit{the third academic term}, \textit{the third semester}, \textit{the three academic terms}, \textit{three terms}, and \textit{trimesters} (see Table~\ref{tab:search_strategy}). We collected 3,000 tweets spanning the period from January 20, 2023 to February 8, 2023.

Tweets for the \textit{Women Driving} target were collected from X using the Octoparse Twitter Scraper. Data were retrieved through Twitter's advanced search interface by restricting the search to Arabic-language tweets and a predefined search period from January 1, 2016 to February 16, 2026. The search query was manually designed using a set of Arabic hashtags and lexical variations related to women driving. It included the following expressions: \textit{you will not drive}, \textit{saudi women driving}, \textit{saudi women drive}, \textit{saudi women driving cars}, \textit{allow women to drive}, \textit{the king allows women to drive}, and \textit{the king supports women driving}. These hashtags were selected based on their active usage during the period covered by the collection, allowing the retrieval of tweets reflecting different perspectives on the target. To reduce irrelevant results, exclusion terms were also added to filter advertisements, promotional content, news headlines, and delivery-related tweets that matched the search keywords. The search strategy yielded an initial collection of 8,450 tweets spanning from January 2, 2016 to November 27, 2025. These tweets were then manually reviewed and processed through the filtering and cleaning step to remove noisy tweets before the annotation process, as described in Section~\ref{Data_cleaning}.

\begin{table*}[htbp]
\centering
\caption{Search strategy used for collecting tweets for each target.}
\includegraphics[width=\textwidth]{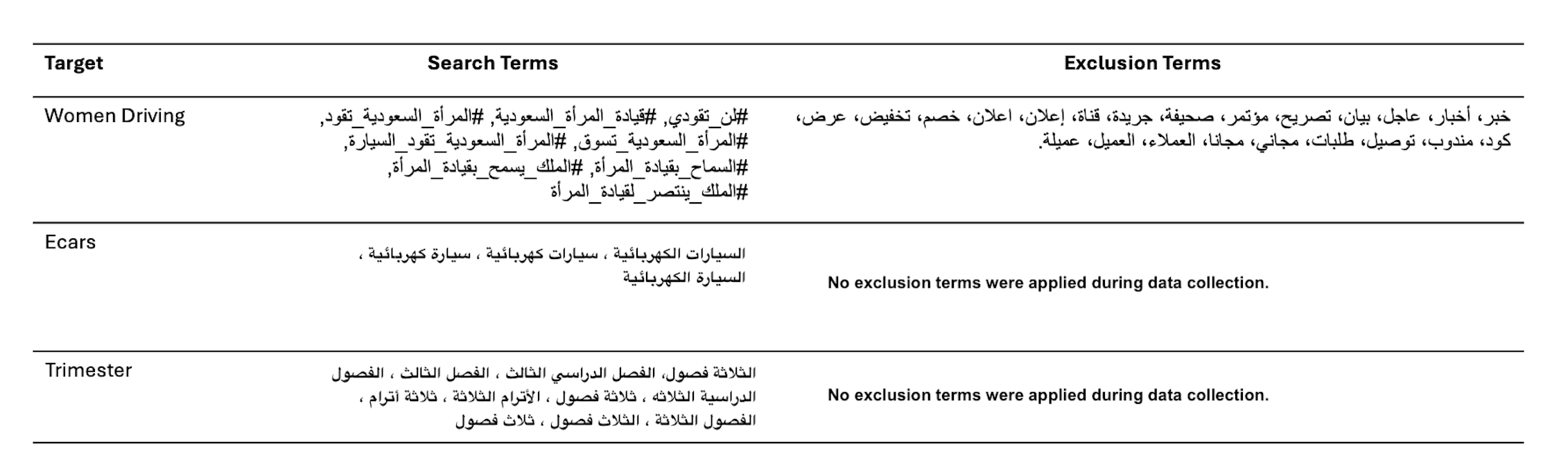}
\label{tab:search_strategy}
\end{table*}

\subsection{Data Cleaning and Filtering}\label{Data_cleaning}

The collected tweets were processed through a series of automatic filtering and preprocessing steps adapted from the original Mawqif dataset. First, only Arabic-language tweets were retained, while tweets written in other languages were removed. Duplicate tweets were then eliminated to ensure unique records. Tweets posted by news and media accounts were identified using the account profile information and excluded from the collection. For the \textit{Trimester System} target, educational accounts were additionally removed to reduce institutional announcements and official communications.

To further improve data quality, a manually constructed list of keywords and phrases was used to filter advertisements and adult-related tweets. For the \textit{Women Driving} target, additional keywords related to driving schools and driving training services were incorporated to remove commercial posts that were not relevant to public opinion toward the target. After the filtering stage, tweets were preprocessed by replacing URLs and user mentions with the special tokens \texttt{URL} and \texttt{MENTION}, respectively, before the annotation stage.

Following automatic filtering, all retained tweets underwent manual quality review to verify their relevance to the target and remove any remaining off-topic or low-quality tweets. For the \textit{Women Driving} target, a random sample of 500 tweets was manually inspected, from which 352 tweets were retained for annotation. For the remaining targets, the selected subsets (5,006 tweets for \textit{E-Cars} and 3,000 tweets for the \textit{Trimester System}) were manually reviewed and filtered before annotation, resulting in 332 \textit{E-Cars} tweets and 312 \textit{Trimester System} tweets.

\subsection{Data Annotation and Verification}\label{Data_Annotation}
The retained tweets were manually following the annotation guidelines of the original Mawqif dataset. Each tweet was assigned three labels: stance (Favor, Against, or None), sentiment (Positive, Negative, or Neutral), and sarcasm (Yes or No). Table~\ref{tab:annotation_labels} summarizes the annotation labels and their definitions. The three annotation tasks were performed independently. For stance annotation, annotators identified the author's intended stance towards the specified target based on both explicit and implicit evidence. The target was explicitly provided during
annotation. Sentiment was annotated independently of stance according to the emotional polarity expressed in the tweet, regardless of the
author's stance towards the target. while sarcasm was annotated based on whether the intended meaning differed from the
literal wording.

\begin{table}[htbp]
\centering
\caption{Annotation labels used in Mawqif-XT.}
\label{tab:annotation_labels}
\small
\begin{tabular}{lll}
\toprule
\textbf{Task} & \textbf{Label} & \textbf{Definition} \\
\midrule

\multirow{3}{*}{Stance}
& Favor & The author expresses or implicitly reveals support for the target. \\
& Against & The author expresses or implicitly reveals opposition to the target. \\
& None & No evidence of support or opposition towards the target, or the tweet is irrelevant to the target. \\

\midrule

\multirow{3}{*}{Sentiment}
& Positive & Expresses a positive opinion or emotional state. \\
& Negative & Expresses a negative opinion or emotional state. \\
& Neutral & Expresses neither positive nor negative sentiment. \\

\midrule

\multirow{2}{*}{Sarcasm}
& Sarcastic & The intended meaning differs from the literal wording, typically expressing criticism or mockery. \\
& Non-sarcastic & The literal meaning is consistent with the intended meaning. \\

\bottomrule
\end{tabular}
\end{table}

Each tweet was independently annotated by three annotators. When disagreements occurred, a fourth annotator reviewed the annotations and assigned the final label according to the annotation guidelines. Inter-annotator agreement was measured using Fleiss' Kappa, with the results reported in Table~\ref{tab:annotation_agreement}.  The obtained agreement scores were 0.564 for stance, 0.485 for sentiment, and 0.526 for sarcasm, indicating moderate agreement for all three annotation tasks according to the interpretation of \cite{landis1977measurement}. Full agreement among all three annotators was observed in 62.95\% of the stance annotations, 54.22\% of the sentiment annotations, and 80.42\% of the sarcasm annotations. At least two annotators agreed on 95.58\% of the stance annotations and 93.98\% of the sentiment annotations.

Further examination of the disagreement cases showed that most
disagreements involved the neutral labels. For stance, 66.12\% of
the disagreement cases involved \textit{None} versus either
\textit{Favor} or \textit{Against}. Similarly, 67.76\% of the
sentiment disagreements involved \textit{Neutral} versus either
\textit{Positive} or \textit{Negative}. This concentration of
disagreements around the neutral categories helps explain the
moderate agreement scores, as annotators differed mainly in deciding
whether a tweet expressed a sufficiently clear stance or sentiment
to receive a polar label or should instead be assigned to the neutral
category. In contrast, direct disagreement between the two polar
labels was less frequent.

\begin{table}[htbp]
\centering
\caption{Inter-annotator agreement measured using Fleiss' Kappa.}
\label{tab:annotation_agreement}
\begin{tabular}{lcccc}
\toprule
\textbf{Annotation} &
\textbf{Women Driving} &
\textbf{Trimester System} &
\textbf{E-Cars} &
\textbf{Overall} \\
\midrule
Stance    & 0.479 & 0.711 & 0.479 & \textbf{0.564} \\
Sentiment & 0.346 & 0.596 & 0.489 & \textbf{0.485} \\
Sarcasm   & 0.486 & 0.469 & 0.541 & \textbf{0.526} \\
\bottomrule
\end{tabular}
\end{table}

\subsection{Dataset Statistics}

 Table~\ref{tab:target_stats} reports the number of tweets for each target and dataset split, while Table~\ref{tab:variables} lists the variables included in the dataset. Table~\ref{tab:descriptive_statistics} summarizes the textual characteristics of tweets across the three targets, including tweet and word lengths, as well as the use of hashtags, mentions, and URLs. While the overall textual characteristics are comparable across targets, slight variations can be observed in average tweet length and the frequency of hashtags, mentions, and URLs.

\begin{figure*}[htbp]
    \centering
    \includegraphics[width=\linewidth]{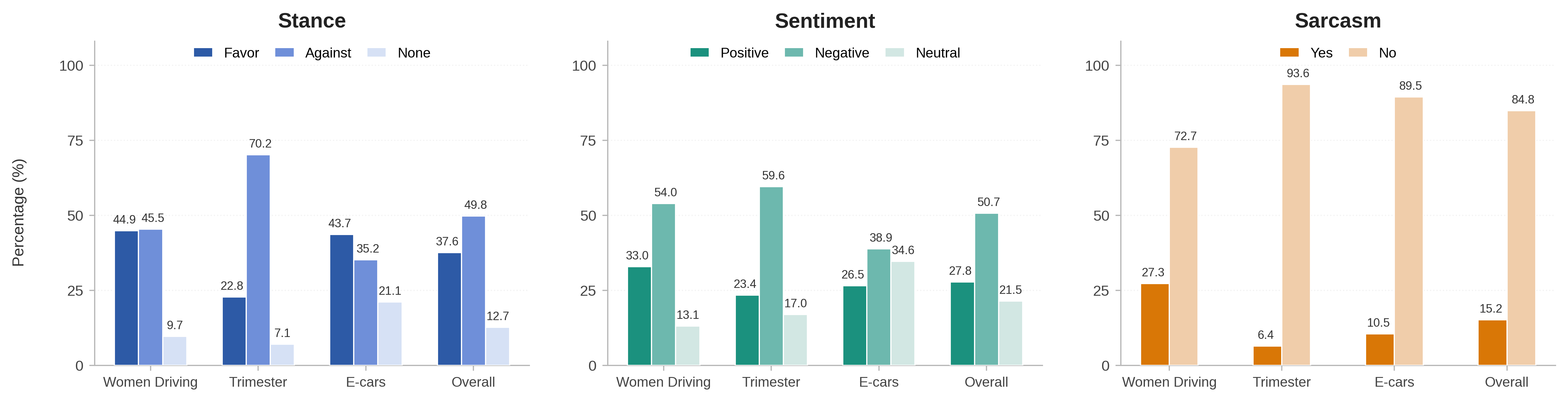}
    \caption{Overall and target-level distributions of stance, sentiment, and sarcasm labels in the Mawqif-XT Extension.}
    \label{fig:target_distribution}
\end{figure*}

The dataset includes two evaluation protocols for cross-target stance detection, as summarized in Table~\ref{tab:protocols}. In the first protocol, models are evaluated on Women Driving, a target that is semantically related to Women Empowerment. In the second protocol, models are evaluated on the unseen E-Cars and Trimester System targets, which belong to different domains. Figure~\ref{fig:target_distribution} show the distributions of stance, sentiment, and sarcasm labels in the new targets.

\begin{table}[htbp]
\centering
\caption{Statistics of the targets included in the Mawqif-XT Extension.}
\label{tab:target_stats}

\begin{tabular}{lccccc}
\toprule
\textbf{Target} &
\textbf{Tweets} &
\textbf{Favor (\%)} &
\textbf{Against (\%)} &
\textbf{None (\%)} &
\textbf{Split} \\
\midrule

Women Driving
& 352 & 44.89 & 45.45 & 9.66 & Test \\

Trimester System
& 312 & 22.76 & 70.19 & 7.05 & Test \\

E-Cars
& 332 & 43.67 & 35.24 & 21.08 & Test \\

\bottomrule
\end{tabular}
\end{table}

\begin{table}[htbp]
\centering
\caption{Description of the variables included in the Mawqif-XT Extension.}
\label{tab:variables}
\begin{tabular}{lp{10cm}}
\toprule
\textbf{Variable} & \textbf{Description} \\
\midrule
id & Unique record identifier. \\
tweet\_id & Original tweet identifier from X. \\
target & Target discussed in the tweet (Women Driving, E-Cars, or Trimester System). \\
stance & Stance label (Favor, Against, or None). \\
sentiment & Sentiment label (Positive, Negative, or Neutral). \\
sarcasm & Sarcasm label (Yes or No). \\

\bottomrule
\end{tabular}
\end{table}

\begin{table}[t]
\centering
\caption{Descriptive statistics of the targets in the Mawqif-XT.}
\label{tab:descriptive_statistics}
\small
\begin{tabular}{lrrrrr}
\toprule
\textbf{Target} & \textbf{Tweets} & \textbf{Avg. No. Words} & \textbf{Avg. No. Hashtags} & \textbf{Avg. No. Mentions} & \textbf{Avg. No. URLs} \\
\midrule

Women Driving          & 352  & 15.75 & 1.21 & 0.01 & 0.03 \\
E-Cars                 & 332  & 26.64 & 0.12 & 0.90 & 0.28 \\
Trimester System       & 312  & 19.74 & 0.31 & 0.77 & 0.07 \\

\bottomrule
\end{tabular}
\end{table}

\begin{table*}[htbp]
\centering
\caption{Evaluation protocols used in the baseline experiments.}
\label{tab:protocols}
\small
\renewcommand{\arraystretch}{1.2}

\begin{tabular}{p{2cm}p{2.5cm}p{2.5cm}p{4cm}p{3.8cm}}
\toprule
&
\multicolumn{2}{c}{\textbf{Original Mawqif}} &
\textbf{Mawqif-XT} &
\\
\cmidrule(lr){2-3}\cmidrule(lr){4-4}
\textbf{Protocol} &
\textbf{Training} &
\textbf{Development} &
\textbf{Test} &
\textbf{Evaluation Setting}
\\
\midrule

Protocol 1
&
\makecell[l]{CV, DT, WE\\3,502 tweets}
&
\makecell[l]{CV, DT, WE\\619 tweets}
&
\makecell[l]{Women Driving\\352 tweets}
&
Related-target generalization
\\

\midrule

Protocol 2
&
\makecell[l]{CV, DT\\2,721 tweets}
&
\makecell[l]{WE\\1400 tweets}
&
\makecell[l]{Trimester System, E-Cars\\644 tweets}
&
Unseen-target generalization
\\

\bottomrule
\end{tabular}

\vspace{1mm}
\footnotesize
CV = COVID-19 Vaccine, DT = Digital Transformation, and WE = Women Empowerment.
\end{table*}

Tables~\ref{tab:example_tweets2} and ~\ref{tab:example_tweets}  present example tweets from the three newly introduced evaluation targets together with their stance, sentiment, and sarcasm labels. Table~\ref{tab:example_tweets2} shows examples where stance and sentiment are aligned, whereas Table~\ref{tab:example_tweets} presents cases where they are not.

\begin{table*}[t]
\centering
\caption{Examples of aligned stance–sentiment polarity in Mawqif-XT.}
\includegraphics[width=\textwidth]{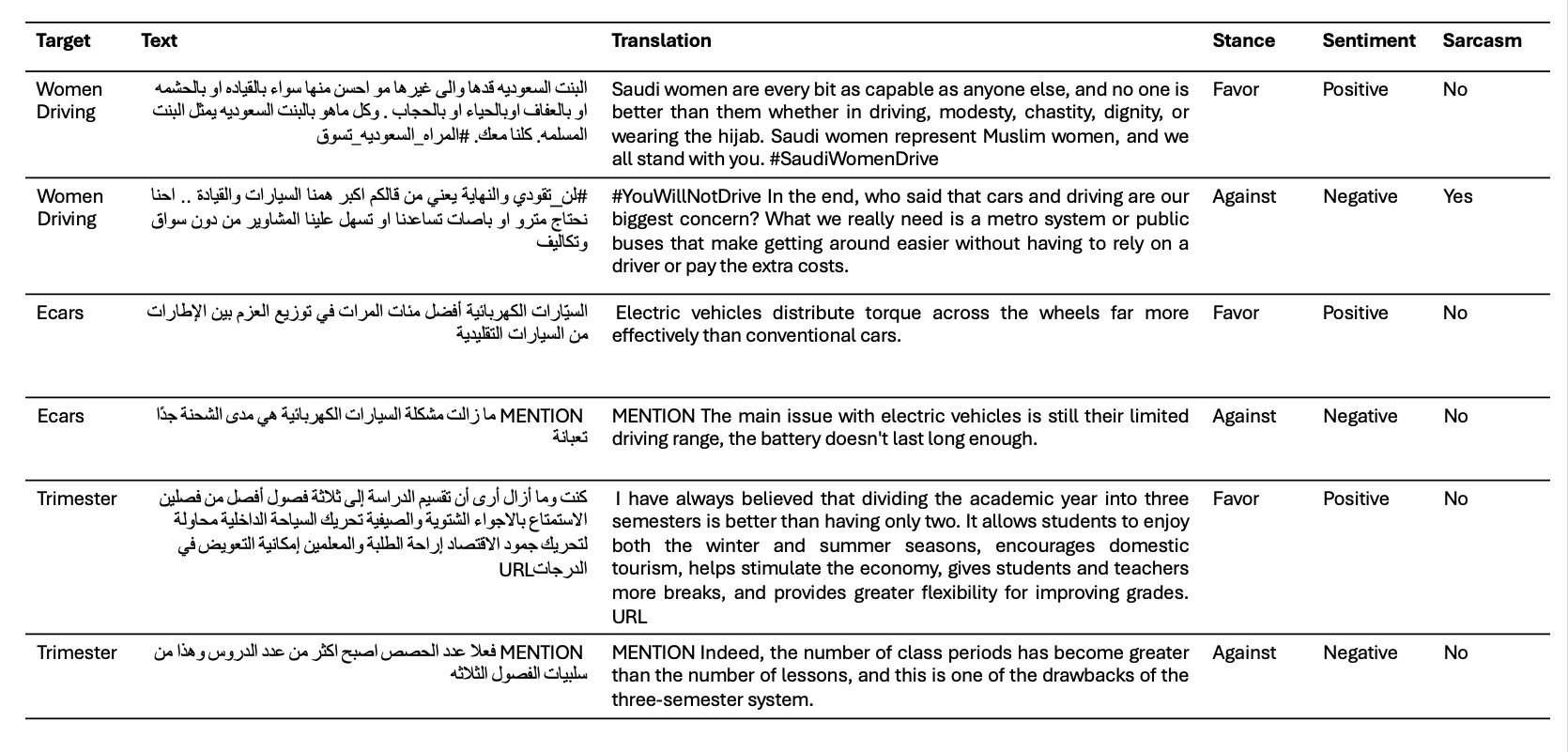}

\label{tab:example_tweets2}
\end{table*}

\begin{table*}[t]
\centering
\caption{Examples of non-aligned stance–sentiment polarity in Mawqif-XT.}
\includegraphics[width=\textwidth]{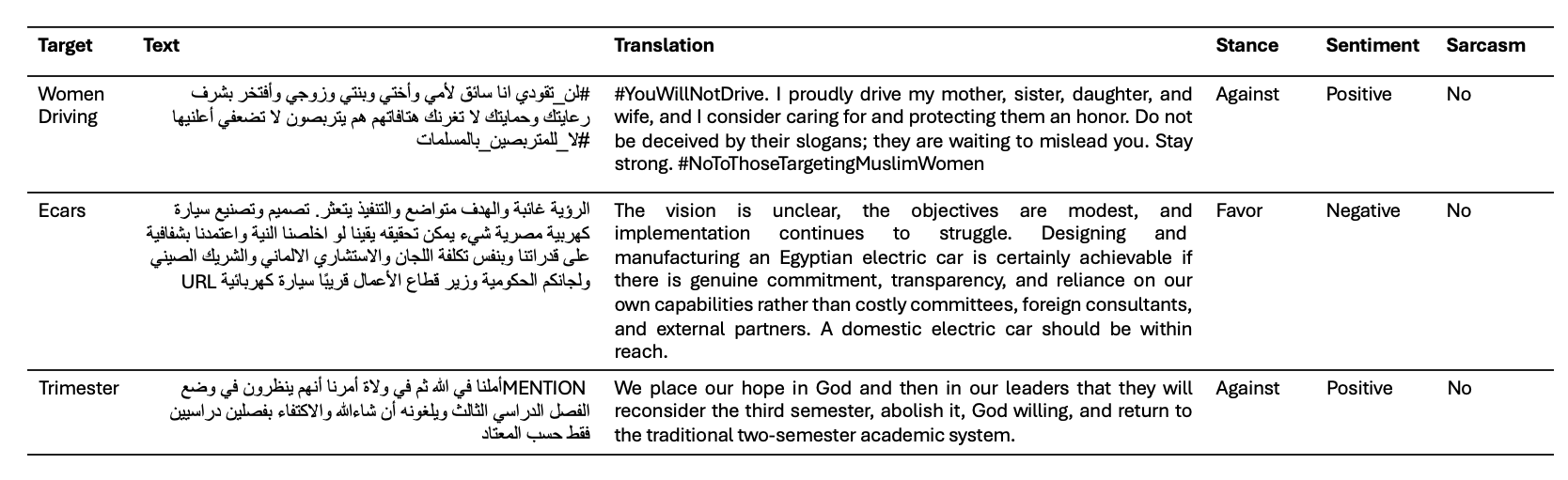}

\label{tab:example_tweets}
\end{table*}

\begin{figure*}[htbp]
    \centering
    \includegraphics[
        width=\textwidth
    ]{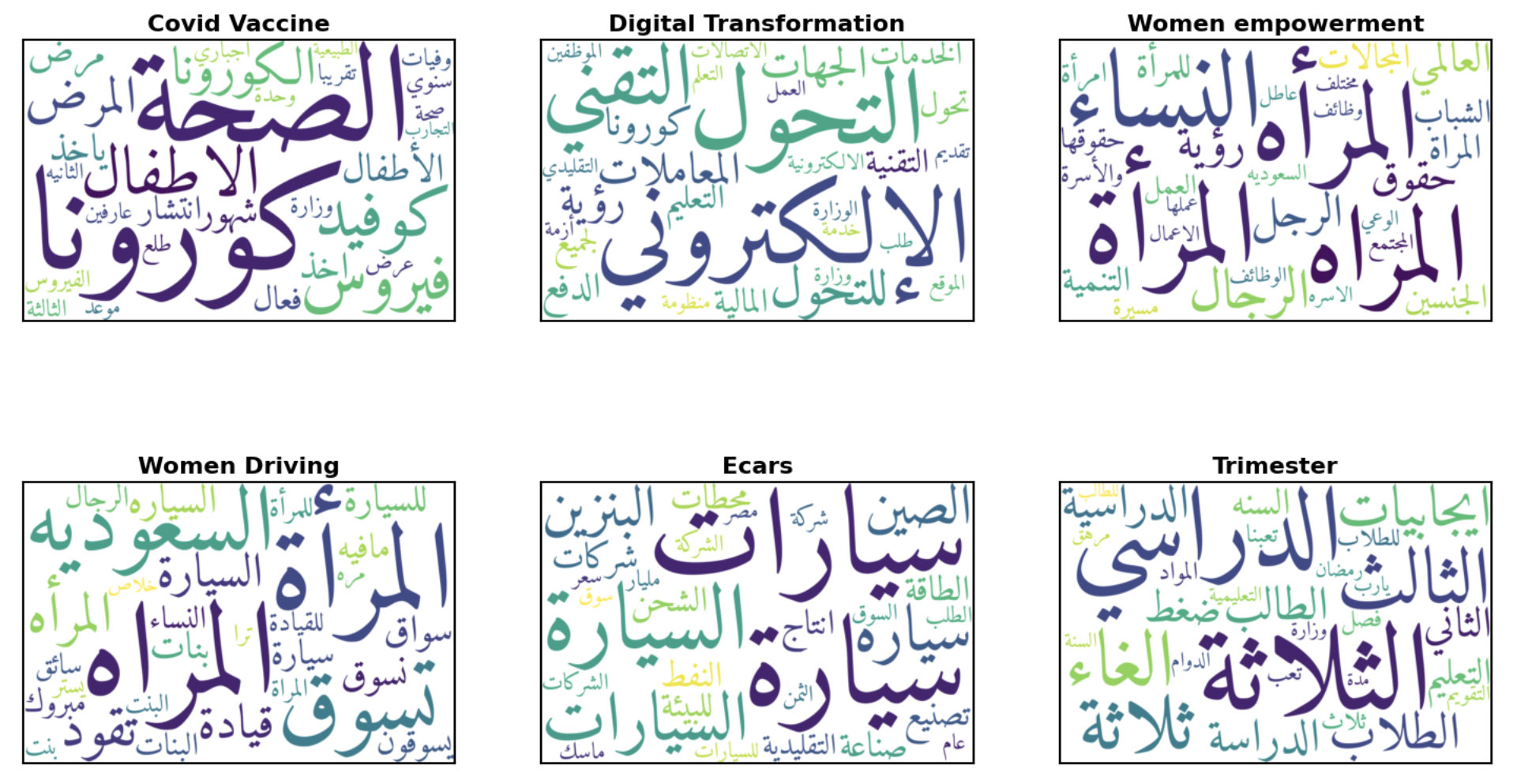}
    \caption{
    TF-IDF-based word clouds comparing the three original Mawqif training targets and the three released evaluation targets.}
    \label{fig:target_wordclouds}
\end{figure*}

To provide a qualitative comparison between the original Mawqif training targets and the released evaluation targets, Figure~\ref{fig:target_wordclouds} presents TF-IDF-based word clouds for all six targets. The word clouds highlight the most representative Arabic terms for each target. Women Driving shares several women-related and social terms with Women Empowerment, whereas E-Cars and Trimester System exhibit distinct automotive- and education-related vocabularies, respectively.

\section{Baseline Experiments}
We evaluate several transformer-based models using the original Mawqif dataset for training and development and the Mawqif-XT Extension for testing. These baselines establish baseline results for future studies and provide a common benchmark for comparing different approaches on the dataset.

\subsection{Experimental Setup}

\subsubsection{Models}
We evaluated four Arabic pretrained language models, namely
AraBERT-v02~\cite{Antoun2020AraBERT},
AraBERT-twitter~\cite{Antoun2020AraBERT},
CAMeLBERT-da~\cite{inoue2021interplay},
and MARBERT~\cite{AbdulMageed2020}. We also included three multilingual models:
mBERT~\cite{DBLP:journals/corr/abs-1810-04805},
DistilBERT~\cite{sanh2019distilbert},
and XLM-RoBERTa~\cite{conneau2020unsupervised}. We also evaluated three large language models (LLMs) in a zero-shot setting: LLaMA 3.3 (70B)~\cite{touvron2023llama}, Qwen 2.5 (72B)~\cite{qwen2.5}, and JAIS (70B)~\cite{sengupta2023jais}. 

\subsubsection{Training Configuration}
All BERT-based models were trained using the same configuration. We applied standard text normalization by removing diacritics, tatweel, non-Arabic characters, and repeated characters. The target and tweet were encoded as a sentence pair. The maximum sequence length was set to 128 tokens, and the batch size was 32. Models were fine-tuned for 20 epochs using the AdamW optimizer with a learning rate of $2\times10^{-5}$. The checkpoint with the best performance on the development set was selected for evaluation. For the zero-shot experiments, we evaluated the three LLMs models using the same inference configuration. Each model received the target and tweet as input and was instructed to predict one of the three stance labels (Favor, Against, or None).

\subsubsection{Evaluation Protocol}
The Mawqif-XT extension supports two evaluation protocols designed to assess cross-target generalization under different transfer settings, as summarized in Table~\ref{tab:protocols}.

\textbf{Protocol 1 (Related-target generalization).}
Models are trained and validated using the original Mawqif training and development splits, which include the COVID-19 Vaccine, Digital Transformation, and Women Empowerment targets. The models are then evaluated on Women Driving, a new target that is semantically related to Women Empowerment. This protocol evaluates the ability of models to transfer to a related target.

\textbf{Protocol 2 (Unseen-target generalization).}
Models are trained on the COVID-19 Vaccine and Digital Transformation targets, validated on Women Empowerment, and evaluated on two unseen targets from different domains: Trimester System and E-Cars. This protocol measures model generalization to distinct unseen targets.
 
We report $F_{avg2}$ as the primary evaluation metric. $F_{avg2}$ is computed as the macro-average F1 score over the Favor and Against classes. Results are reported separately for each target. An overall score is also reported for each evaluation setting by computing $F_{avg2}$ over all test instances in that setting.

\subsection{Results}

Table~\ref{tab:Results-all} summarizes the stance detection performance of all evaluated models under the two evaluation protocols. Under Protocol~1, models were trained and validated on the original Mawqif targets and evaluated on the related Women Driving target. Across all models, Qwen~2.5 (72B) achieved the highest development score on COVID-19 Vaccine ($F_{avg2}=84.70$) and the highest overall development score ($F_{avg2}=84.25$). AraBERT-twitter obtained the best result on Digital Transformation ($F_{avg2}=79.55$), while JAIS (70B) achieved the highest score on Women Empowerment ($F_{avg2}=87.20$). On the Women Driving test set, Qwen~2.5 achieved the best performance ($F_{avg2}=73.66$). Among the encoder-based models, AraBERT-twitter achieved the highest overall development score ($F_{avg2}=83.90$), whereas AraBERT-v02 obtained the best Women Driving test result ($F_{avg2}=65.92$).

Under Protocol~2, models were trained on the COVID-19 Vaccine and Digital Transformation targets, validated on Women Empowerment, and evaluated on the unseen E-Cars and Trimester System targets. JAIS achieved the highest development score on Women Empowerment ($F_{avg2}=87.20$). On the test set, Qwen~2.5 obtained the best result on E-Cars ($F_{avg2}=80.60$) and the highest overall score ($F_{avg2}=74.75$), while MARBERT achieved the highest performance on the Trimester System target ($F_{avg2}=74.11$). Among the encoder-based models, MARBERT obtained the highest overall test score ($F_{avg2}=72.21$).

Overall, the results show that no single model achieved the highest performance across all targets. Although Qwen~2.5 obtained the strongest overall results in both protocols, AraBERT-twitter, JAIS, and MARBERT achieved the best performance on specific targets. This variation highlights differences across the evaluation targets and provides baseline results for studying transfer to related targets and generalization to unseen domains. These baseline results provide a usefull reference for future work on Arabic cross-target stance detection.

\begin{table*}[htbp]
\centering
\caption{
Stance detection results on Mawqif-XT under the two evaluation protocols. \small\textit{CV: COVID-19 Vaccine, DT: Digital Transformation, WE: Women Empowerment, WD: Women Driving, EC: E-Cars, and TR: Trimester. Best results in each column are shown in bold.}
}

\label{tab:Results-all}

\resizebox{\textwidth}{!}{%
\begin{tabular}{@{}lccccc|cccc@{}}
\toprule

& \multicolumn{5}{c|}{\textbf{Protocol 1: Related Target}}
& \multicolumn{4}{c}{\textbf{Protocol 2: Unseen Targets}} \\

\cmidrule(lr){2-6}
\cmidrule(lr){7-10}

& \multicolumn{4}{c}{\textbf{Development Set}}
& \multicolumn{1}{c|}{\textbf{Test Set}}
& \multicolumn{1}{c}{\textbf{Development Set}}
& \multicolumn{3}{c}{\textbf{Test Set}} \\

\cmidrule(lr){2-5}
\cmidrule(lr){6-6}
\cmidrule(lr){7-7}
\cmidrule(lr){8-10}

\textbf{Model}
& CV
& DT
& WE
& Overall
& WD
& WE
& EC
& TR
& Overall \\

\midrule

\multicolumn{10}{@{}l}{\textbf{Arabic BERT Models}} \\

CAMeLBERT-da
& 75.20 & 65.64 & 79.88 & 77.77 & 62.49
& 75.42 & 61.47 & 62.56 & 62.65 \\

MARBERT
& 79.82 & 73.90 & 84.73 & 81.91 & 64.53
& 79.45 & 68.05 & \textbf{74.11} & 72.21 \\

AraBERT-v02
& 79.61 & 76.00 & 85.11 & 82.14 & 65.92
& 78.86 & 65.58 & 68.75 & 67.82 \\

AraBERT-twitter
& 81.39 & \textbf{79.55} & 85.29 & 83.90 & 64.84
& 79.88 & 58.00 & 63.95 & 62.77 \\

\midrule

\multicolumn{10}{@{}l}{\textbf{Multilingual Models}} \\

mBERT
& 74.19 & 67.62 & 76.58 & 76.11 & 61.69
& 63.63 & 54.61 & 52.06 & 54.18 \\

DistilBERT-multilingual
& 69.47 & 66.07 & 74.84 & 73.57 & 56.43
& 63.41 & 55.68 & 52.42 & 55.21 \\

XLM-R-base
& 71.03 & 68.18 & 82.16 & 76.83 & 60.39
& 72.22 & 61.15 & 51.02 & 56.49 \\

\midrule

\multicolumn{10}{@{}l}{\textbf{LLM-based Models (Zero-shot)}} \\

LLaMA 3.3 (70B)
& 79.87 & 73.61 & 78.96 & 79.41 & 67.08
& 78.96 & 67.40 & 61.46 & 66.39 \\

Qwen 2.5 (72B)
& \textbf{84.70} & 73.38 & 86.34 & \textbf{84.25} & \textbf{73.66}
& 86.34 & \textbf{80.60} & 68.44 & \textbf{74.75} \\

JAIS (70B)
& 82.29 & 71.53 & \textbf{87.20} & 83.37 & 72.67
& \textbf{87.20} & 71.26 & 69.76 & 69.10 \\

\bottomrule
\end{tabular}%
}
\end{table*}

\section{Usage Notes}
\label{sec:usage}

Mawqif-XT dataset can be used together with the original Mawqif dataset for Arabic stance detection as well as related tasks such as sentiment analysis, sarcasm detection, and multi-task learning. 
The released evaluation dataset supports two evaluation protocols for cross-target stance detection. The original Mawqif dataset is used for training and development, whereas the released dataset is reserved for testing. Women Driving enables evaluation on a semantically related target, while E-Cars and Trimester System evaluate model generalization to previously unseen targets from different domains. This setup provides a standardized benchmark for evaluating cross-target generalization in Arabic stance detection. Since each tweet is annotated with stance, sentiment, and sarcasm labels, the dataset can also support research on the relationship between these tasks in both single-task and multi-task learning settings.

\section{Limitations}\label{sec:limitations}

Mawqif-XT has several limitations. First, although the dataset extends the original Mawqif benchmark with three additional public targets, these targets cover only a limited range of domains. As a result, the dataset does not represent the full diversity of topics discussed in Arabic social media. Second, the new targets include one target that is semantically related to the training data and two targets from different domains. While this setting supports the evaluation of cross-target generalization, it does not cover every possible transfer scenario. Finally, the dataset was collected from X (formerly Twitter), and therefore reflects the language and user interactions on that platform during the data collection period.

\section{Conclusion}\label{sec:Conclusions}
This paper presents the Mawqif-XT Extension, which adds 996 manually annotated Arabic tweets covering three new targets: \textit{Women Driving}, \textit{E-Cars}, and \textit{Trimester System}. The extension follows the annotation scheme of the original Mawqif dataset and provides annotations for stance, sentiment, and sarcasm. The new targets extend the available data to support the study and evaluation of Arabic stance detection across different targets. We hope that this resource will support future research on Arabic stance detection and related NLP tasks.

%\section*{Ethics Statement}
%[Mandatory: State compliance with ethical standards if human/animal trials were involved, or explicitly state: "Not applicable."]

%\section*{Declaration of Competing Interest}
%The authors declare that they have no known competing financial interests or personal relationships that could have appeared to influence the work reported in this paper.

%\section*{CRediT Author Statement}
%\textbf{Author One:} Conceptualization, Methodology, Data curation. \textbf{Author Two:} Writing - review \& editing.

\section*{Acknowledgments}
The authors would like to acknowledge the support provided by the Deanship of Research (DR) of King Fahd University of Petroleum \& Minerals (KFUPM) for funding
this work through project No. EC2620.

%% REFERENCES
\bibliographystyle{elsarticle-num} 
\bibliography{cas-refs}

\end{document}